# DSTP-RNN: a dual-stage two-phase attention-based recurrent neural networks for long-term and multivariate time series prediction

Yeqi Liu[1,2,3], Chuanyang Gong[1,2,3], Ling Yang[1,2,3], Yingyi Chen[1,2,3,*]

**Abstract**

Long-term prediction of multivariate time series is still an important but challenging problem. The key to solve this problem is to capture the spatial correlations at the same time, the spatio-temporal relationships at different times and the long-term dependence of the temporal relationships between different series. Attention-based recurrent neural networks (RNN) can effectively represent the dynamic spatio-temporal relationships between exogenous series and target series, but it only performs well in one-step time prediction and short-term time prediction. In this paper, inspired by human attention mechanism including the dual-stage two-phase (DSTP) model and the influence mechanism of target information and non-target information, we propose DSTP-based RNN (DSTP-RNN) and DSTP-RNN-Ⅱ respectively for long-term time series prediction. Specifically, we first propose the DSTP-based structure to enhance the spatial correlations between exogenous series. The first phase produces violent but decentralized response weight, while the second phase leads to stationary and concentrated response weight. Secondly, we employ multiple attentions on target series to boost the long-term dependence. Finally, we study the performance of deep spatial attention mechanism and provide experiment and interpretation. Our methods outperform nine baseline methods on four datasets in the fields of energy, finance, environment and medicine, respectively.

**Keywords:** Time series prediction; Spatio-temporal relationship; Attention mechanism; Dual-stage two-phase model; Deep attention network.

**1 Introduction**

Recent developments in the Internet of Things and Big Data have led to the continuous expansion of data scale(Le & Ge, 2019). Similarly, time series are increasingly showing multivariate feature in the field of energy consumption forecasting(Luis M. Candanedo, Veronique Feldheim, & Deramaix, 2017), financial market prediction(Moews, Herrmann, & Ibikunle, 2018; Qin, Song, Cheng, Cheng, & Cottrell, 2017), environment forecasting(Zamoramart ńez, Romeu, Botellarocamora, & Pardo, 2014), heart and brain signal analysis(S. M., M. A., J. C., & J. M., 2018), etc. Hence, the long-term prediction of multivariate time series has more practical significance, e.g., it is more significant to forecast the weather of one or more days than to forecast the weather of the next hour in the future. However, the long-term prediction of multivariate time series is still a challenging problem, which is mainly reflected in the feature representation and selection mechanism of spatio-temporal relationships between different series. Specifically, it is embodied in three aspects (Figure 1(c)): the spatial correlations between different attributes at the same time, the spatio-temporal relationships between different attributes at different times, and the temporal relationships between different series(Monidipa & Ghosh; Qin, et al., 2017; Yunzhe, et al., 2018).

Although the time series prediction has attracted a wide attention of the community, typical methods, such as autoregressive integrated moving average model (ARIMA)(Amini, Kargarian, & Karabasoglu, 2016), kernel method(Jie & Zio, 2016), and recurrent neural networks (RNN)(Chen, Xin, She, & Min, 2017), mainly focus on solving one aspect of the dynamic spatio-temporal relationships, so it is impossible to achieve accurate and robust long-term prediction of multivariate time series. However, since the attention mechanism can learn the weight distribution strategy of raw data, and further enhance the feature representation ability of spatio-temporal relationships, attention-based RNN have been used to understand the spatial correlations in time series prediction (Edward Choi & Stewart., 2016). Recently, Qin et al.(Qin, et al., 2017) and Liang et al.(Yuxuan, Songyu, Junbo, Xiuwen, & Yu, 2018) combined attention mechanism and encoder-decoder model to achieve state-of-the-art performance in single step and short-term time series prediction, which have done inspiring works in the fields of multi-variable time series forecasting and Geo-sensory time series forecasting, respectively. Therefore, attention-based encoder-decoder methods have shown their success to extract the spatio-temporal dependence of multivariate time series. Nevertheless, these methods only perform well in one-step time prediction and short-term time prediction.

In addition to the attention-based RNN, our work is also inspired by DSTP model of human attention and the target and non-target information mechanism of human neuron signals(Ronald, Marco, & Carola, 2010). On the one hand, in the first stage of DSTP model, the response has been selected in the first phase leads to violent but decentralized response, but it will produce stationary and concentrated response in the second stage(Ronald, et al., 2010). The attention mechanism in different stages can be embodied in the two stages including spatial attention and temporal attention when we design artificial neural network structure, and the attention mechanism of different phases is reflected in the multiple filtering of spatial correlations in the first stage. Hence, we propose DSTP-RNN model with dual-stage two-phase structure. On the other hand, the stimulation of neuron signal shows that both target signal and non-target signal have some effect, because perceptual filtering is imperfect(Ronald, et al., 2010). In fact, the supervised dataset reconstruction based on target series is the key to use traditional machine learning method for time series prediction, which reflects the effect of past information of target series on forecasting. Meanwhile, the RNN method, which forecasts future values based on past values of time series, shows that the future temporal dependence of time series depends on its own past values. Therefore, we develop DSTP-RNN-II model and our network structure pays more attention to the spatio-temporal relationships between target series and exogenous series. Furthermore, the attention mechanism of human vision is a multi-layer neuron structure, which is widely used in the natural language processing(Vaswani, et al., 2017) and computer vision(Y. Li, Zeng, Shan, & Chen, 2018). Naturally, we study the deep attention mechanism in the spatial attention.

In this paper, we enhance the focus on spatial correlations through the DSTP-based model, and enhance the attention to temporal relationships through the embedding of target information, thus capturing more accurate spatio-temporal relationships. The contributions of our work are four-fold:

- **DSTP-RNN.** Inspired by the DSTP model of human attention, we propose the DSTP-RNN model to capture the spatio-temporal relationships between multivariate time series. Specifically, two phases mean the use of two consecutive phase attention with and without target series to capture spatial correlations, and these phases differ with respect to their susceptibility to interference. Dual stages refer to the spatial attention mechanism for the original series and the temporal attention mechanism for the hidden state in the spatial attention of last phase.
- **Target and no-target information mechanism.** Enlightened by the target and non-target information mechanism of human neuron signals(Zhang, Xiong, & Su, 2018), we develop DSTP-RNN-II to extract the spatial relationships between target series and exogenous series through a parallel spatial attention module. Furthermore, we allocate more attention mechanism on the past value of target information to better obtain the long-term dependence. Specifically, we embedded the values of target series corresponding to exogenous series in dataset reconstruction and last-phase spatial attention mechanism.

- **Deep spatial attention.** Due to the multi-layer structure of human neural network(Fukushima & Miyake, 1982), we further study the effect of deep spatial attention mechanism on spatio-temporal relationships and give the interpretation experiments.
- **Application in many fields.** We demonstrate the effectiveness and applicability of our methods on four open datasets from different fields. Extensive experiments show that our methods can achieve state-of-the-art results.

## 2 Related Work

Our work is mainly related with two lines of research: time series prediction and attention-based neural network structure.

In current classical methods for time series prediction, ARIMA (Amini, et al., 2016; Geetha & Nasira, 2016) only focuses on the seasonality and regularity of target series, which can effectively extract the long-term dependence of the series itself, while ignoring the spatial correlations of exogenous series. In addition, it assumes that series changes are stable, so the ARIMA model and its variants are not suitable for non-stationary and multivariable time prediction. When traditional regression methods based on machine learning, e.g., support vector regression (SVR)(Gestel, et al., 2001; Jie & Zio, 2016), are used to predict time series, the exogenous series are mapped in the high-dimensional space, which pays more attention to the spatial correlations of these exogenous series at the same time, but ignores the time dependence. The discrete spatial distribution in high-dimensional space cannot represent the temporal characteristics of time series. On the contrary, RNN approaches(Han & Xu, 2018; Sivakumar & Sivakumar, 2017) focus on maintaining the continuous time dependence, especially as LSTM(Graves, 1997) and GRU(Chung, Gulcehre, Cho, & Bengio, 2014) give simple linear operation for neuron information and add the external information of the current moment through the gating mechanism. However, the vectorization of simultaneous attributes without attribute selection cannot select spatial features between different series.

Attention-based neural network structure was first used in natural translation(Bahdanau, Cho, & Bengio, 2015), and now it is widely used as an important middle layer in machine translation(Gangi & Federico, 2018), image caption(Yong, et al., 2017), time series prediction(Guo & Lin, 2018; Yuxuan, et al., 2018), etc. Attention networks can be divided into three classes: single-layer attention mechanism(H. Li, Shen, & Zhu, 2018), double-layer attention mechanism and deep attention mechanism(Tixier, 2018). Specifically, the double-layer attention mechanism includes the parallel attention mechanism(Seo, Huang, Yang, & Liu, 2017), the hierarchical attention mechanism (Tixier, 2018; Yang, He, Gao, Deng, & Smola, 2016) and the hybrid attention mechanism(Zhao & Zhang, 2018). In general, different attention networks can be applied to different tasks, but deeper attentions can be trained to obtain more accurate weights(Zhang, et al., 2018). In this paper, we study the hierarchical attention mechanism (DSTP-RNN), the hierarchical and parallel hybrid attention mechanism (DSTP-RNN-Ⅱ) and the deep attention mechanism (DeepAttn) in time series prediction.

Recently, attention-based neural network structure has been applied to multivariate time series prediction, but attention mechanism can only choose the spatial relationships at the same time, ignoring the spatio-temporal relationships at different times(Edward Choi & Stewart., 2016). To solve this problem, since encoder-decoder model maintains the spatio-temporal relationships by encoding the series within a certain time window into a fixed-length vector(Ilya, Oriol, & Quoc, 2014; Nal & Phil,

2013), this method can extract spatial correlations by giving different weights to different exogenous series, and obtains spatio-temporal relationships at different times by encoding the series with a fixed length. Insufficiently, the finite length vectors cannot feedback the long-term spatio-temporal characteristics of different attributes at different times. With the increase of the length of vector representation, the performance of the encoder-decoder network will deteriorate rapidly(Cho, Van, Merriënboer, Bahdanau, & Bengio, 2014). Therefore, Bahdanau et al.(Bahdanau, et al., 2015) proposed the attention mechanism that belongs to encoder-decoder family to address this issue.

Due to the success of attention-based encoder-decoder networks in sequence learning, Qin et al. (Qin, et al., 2017) and Liang et al. (Yuxuan, et al., 2018) employ two-stage attention mechanism to forecast multivariate time series. In the first stage, the spatial attention is used to obtain spatial correlations. In the second stage, the relevant hidden states in spatial stages are selected through the temporal attention mechanism to ensure time dependence. Also, it encodes all hidden states into a context vector to capture spatio-temporal relationships at different times. Hence, these networks take into account the spatial correlations at the same time and at the different times, as well as the temporal dependence in the time dimension.

However, current attention-based RNN methods ignore the spatial correlations between target series and exogenous series. In addition, the weights obtained by the single-layer spatial attention mechanism are scattered and violent, and thus it is only suitable for one step or short-term prediction. Actually, the single phase is not strictly selective that information from irrelevant attributes might also be assigned a large attention weight, which is also susceptible to interference. In comparison, we use two-phase attention mechanism to extract more accurate spatial correlations between target series and exogenous series for long-term prediction. The first phase produces violent but decentralized attention weight, while the second phase leads to stationary and concentrated attention weight. Additionally, more attention mechanisms are allocated to target series to capture the long-term dependence. Finally, we also study the response of deeper spatial attention mechanism to spatial correlations.

## 3 Preliminary

### 3.1 Notation

Given n (n ≥ 1) exogenous series and one target series, we use $\boldsymbol{x^k} = (x_1^k, x_2^k, ..., x_T^k)^T \in R^T$ to represent k-th exogenous series within the length of window size $T$, and we use $\mathbf{X} = (\boldsymbol{x_1}, \boldsymbol{x_2}, ..., \boldsymbol{x_T})^T \in R^{n \times T}$ ($\boldsymbol{x_k} = \boldsymbol{x^k}$) to represent all exogenous variables within window size $T$.

As for the notation related to target series, we use $\mathbf{Y} = (y_1, y_2, ..., y_T)^T \in R^T$ to represent the target series within window size $T$, we use $\mathbf{Z} = (\boldsymbol{z_1}, \boldsymbol{z_2}, ..., \boldsymbol{z_T})^T \in R^{(n+1) \times T}$ to represent the set of the output of first phase attention ($\widetilde{\mathbf{X}}$) and target series ($\mathbf{Y}$), and we use $\widehat{\mathbf{Y}} = (y_{T+1}, y_{T+2}, ..., y_{T+\tau})^T \in R^\tau$ to represent the future values of target series, where $\tau$ is the time step to be predicted.

### 3.2 Problem Statement

Given the previous values of the exogenous series and target series, i.e., $(\boldsymbol{x_1}, \boldsymbol{x_2}, ..., \boldsymbol{x_T})$ with $\boldsymbol{x_t} \in R^n$, and $(y_1, y_2, ..., y_T)$ with $y_t \in R$, we aim to predict the future values over next $\tau$ time steps, denoted as $\widehat{\mathbf{Y}} = (y_{T+1}, y_{T+2}, ..., y_{T+\tau})^T \in R^\tau$, which is shown as follow.

$$\hat{y}_{T+1}, \hat{y}_{T+1}, ..., \hat{y}_{T+\tau} = F(y_1, ..., y_T, \boldsymbol{x_1}, ..., \boldsymbol{x_T}) \quad (1)$$

Where $F(\cdot)$ is a nonlinear mapping function we aim to learn.

## 4 Model

Figure 1(a) and Figure 1(b) present the overall framework of the proposed DSTP-RNN and DSTP-RNN-Ⅱ, respectively. Dual-stage refers to the selection of spatial features in the first stage and temporal features in the second stage, which named spatial attention (red box in Figure 1) and temporal attention (blue red box in Figure 1). The spatial attention module consists of two-phase structure. The first phase produces violent but decentralized response weight from the original data, while the second phase leads to stationary and concentrated response weight from the weighed data of previous phase. Inspired by the human attention model, both target and non-target information affect the response of neuron stimulus, so the DSTP-RNN-Ⅱ model with a parallel attention module is closer to the selection process of human attention model. Recently, deep attention mechanism performs well in machine translation(Zhang, et al., 2018), and we report the results of third layers of spatial attention mechanism, which is named DeepAttn. The deep attention network is a hierarchical depth structure based on Figure 1(a), and it only stacks one spatial attention unit. Note, the input of the last-phase spatial attention in each method concatenates the value of the target series at the corresponding time ($\mathbf{y}^T$), thereby more closely linking the spatio-temporal relationships. Figure 1(c) shows an example of the dual-stage data mapping process within a time window, and the multi-phase model only repeats the first stage. In addition, the attention module including single-layer LSTM structure is used as one of the basic units, and other attention mechanism (Luong, Pham, & Manning, 2015) and RNN structures can be tried(Zhang, Xiong, & Su, 2017).

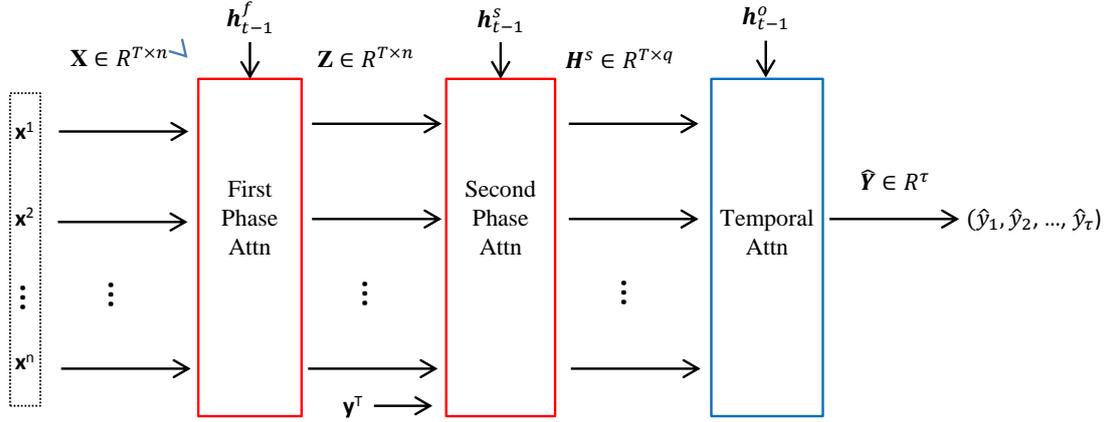

(a) Overall framework of the proposed DSTP-RNN model

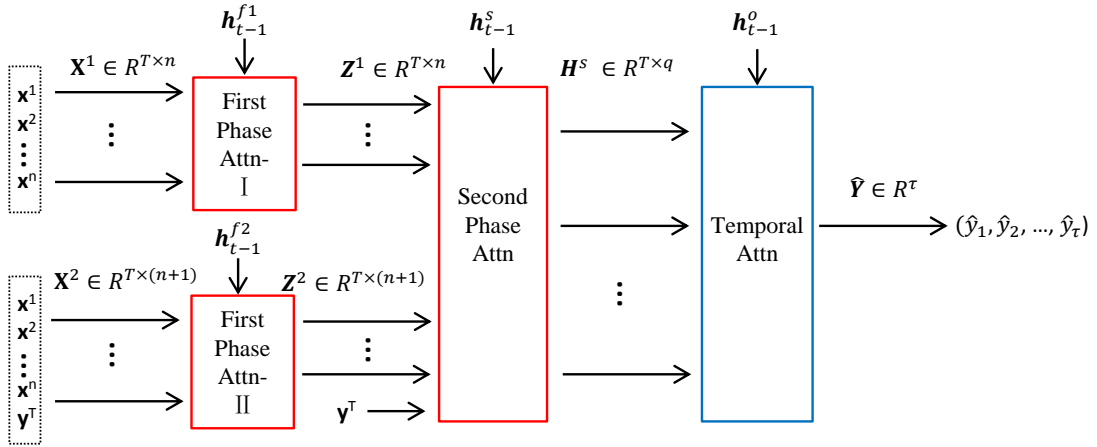

(b) Overall framework of the proposed DSTP-RNN-Ⅱ model

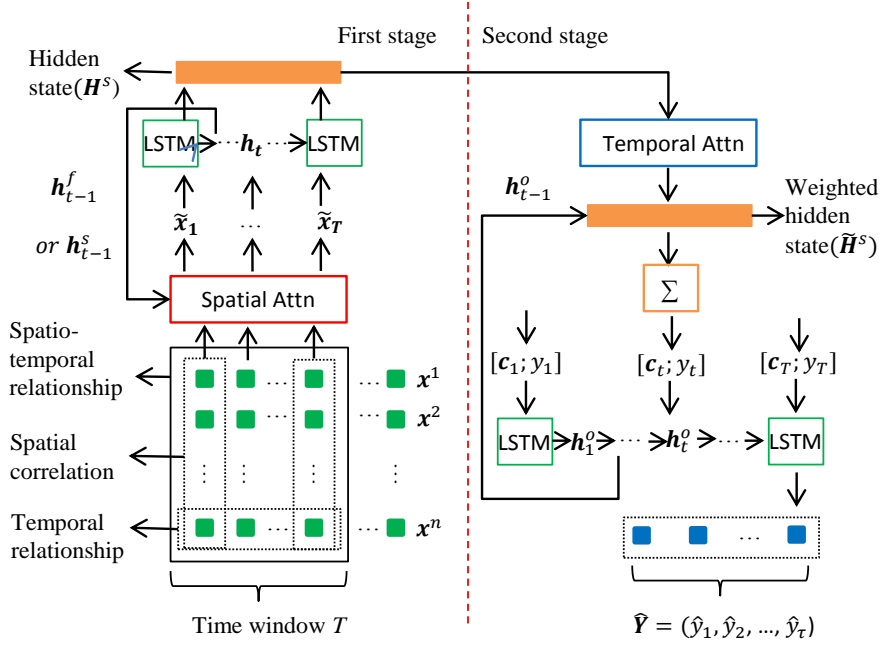

(c) Model details within a time window as an example

Figure 1: Illustration of our approaches. The red box represents the spatial attention with the original series (e.g., $X$), the combined series (e.g., $X^2$), or the weighted series (e.g., $Z$) as input, and the blue box represents the temporal attention based on the weighting operation for the encoder hidden state (e.g., $H^s$). DeepAttn only stacks one spatial attention unit based on DSTP-RNN, in which the weighted data of the second phase attention module is used as input. The notation meaning and details of the vector mapping are presented in Sections 3 and Section 4.

**4.1 Spatial attention**

The attention mechanism of the first and second phases belongs to the spatial attention mechanism. The purpose is to extract the spatial correlations between exogenous series and target series. Specifically, the first phase is used to extract spatial correlations of exogenous series. The second phase is used to extract weighted features again, and it is also use to extract spatial correlations between exogenous series and target series. Hence, it ensures that the extracted spatial features are stable and efficient.

**4.1.1 First phase attention**

In the first phase, we only employ an attention module for exogenous series to obtain the spatial correlations between these exogenous attributes. Also, due to the influence mechanism of target information and non-target information(Ronald, et al., 2010), a variant is to use an independent attention module to obtain the relationships between target series and exogenous series, which is the idea source for the first phase attention-Ⅱ module in DSTP-RNN-Ⅱ model. In fact, the main difference between these two types of the first phase attention module, i.e., first phase attention-Ⅰ and first phase attention-Ⅱ, is whether the exogenous series is concatenated with the target variable at the corresponding time (i.e., $\ddot{x}_t^k = [x_t^k; y_t^k]$ for the k-th attribute vector at time $t$). Typically, given the k-th attribute vector of any exogenous series at time $t$ (i.e., $x^k$), we can employ the following attention mechanism:

$$f_t^k = v_f^T \tanh(W_f[h_{t-1}^f; s_{t-1}^f] + U_f x^k + b_f) \quad (2)$$

$$\alpha_t^k = \frac{\exp(f_t^k)}{\sum_{j=1}^n \exp(f_t^j)} \quad (3)$$

Where $[*;*]$ is a concentration operation, and $v_f, b_f \in R^T$, $W_f \in R^{T \times 2m}$, $U_f \in R^{T \times T}$ are parameters to learn. Here, $h_{t-1}^f \in R^m$ and $s_{t-1}^f \in R^m$ are respectively the hidden state and cell state of the previous hidden LSTM unit in the encoder, and $m$ is the hidden size in this attention module. The attention weights are determined by historical hidden state and current input in the encoder, which represents the impact of each attribute on forecasting. Since any attribute value at any time has its corresponding weight, the output after the first phase attention weighting is defined as follows:

$$\tilde{x}_t = (\alpha_t^1 x_t^1, \alpha_t^2 x_t^2, \dots, \alpha_t^n x_t^n)^T \quad (4)$$

**4.1.2 Second phase attention**

The input of second phase attention is the vector that concatenates the output of the first phase attention and the value of target series at corresponding time. The more stationary and concentrated spatial correlations can be obtained through this second selection of spatial relationships. In this module, we combine all past values of target series with exogenous series at corresponding time. Note that when we use an independent attention mechanism (i.e., first phase attention-Ⅱ) to obtain spatial relationships between target series and exogenous series, we concatenate the output from these two types of the first phase attention as the input to the second phase attention, which is operated in DSTP-RNN-Ⅱ model. As for input data of this module, we concatenate corresponding target variable $y^k$ to the k-th attribute $\tilde{x}^k$ to construct a vector $z^k$, i.e., $z^k = [\tilde{x}^k; y^k] \in R^{(n+1) \times T}$, and we calculate attention weight as follows:

$$s_t^k = v_s^T \tanh(W_s[h_{t-1}^s; s_{t-1}^s] + U_s[\tilde{x}^k; y^k] + b_s) \quad (5)$$

$$\beta_t^k = \frac{\exp(s_t^k)}{\sum_{j=1}^{n+1} \exp(s_t^j)} \quad (6)$$

Where $v_s, b_s \in R^T$, $W_s \in R^{T \times 2q}$, $U_s \in R^{T \times T}$ are parameters to learn, and $h_{t-1}^s \in R^q$ and $s_{t-1}^s \in R^q$ are respectively the hidden state and cell state of the previous hidden LSTM unit in the encoder, and $q$ is the hidden size in this attention module. For each spatial attention module with target series (e.g., first phase attention-Ⅱ), we will employ the above operation independently. The output after the second phase attention is as follows:

$$\tilde{z}_t = (\beta_t^1 z_t^1, \beta_t^2 z_t^2, \dots, \beta_t^{n+1} z_t^{n+1})^T \quad (7)$$

**4.2 Temporal attention**

The method of maintaining time dependence in spatial attention includes the spatio-temporal relationships in time window $T$ and concatenation vectors between target series and exogenous series, both of which reflect the temporal relationships within a fixed window. Since the temporal relationships in a fixed window are not enough, longer time dependence still needs attention mechanism to select the hidden state of the encoder. Specifically, the long-term dependence of time series can be obtained by selecting the hidden state in the encoder that is most related to the target value to be predicted, which also connects spatio-temporal relationships in this way. For each i-th hidden state from the second phase attention, the temporal relationships can be obtained by attention mechanism as follows:

$$d_t^i = v_d^T \tanh(W_d[h_{t-1}^o; s_{t-1}^o] + U_d h_i^s + b_d) \quad (8)$$

$$\gamma_t^i = \frac{\exp(d_t^i)}{\sum_{j=1}^T \exp(d_t^j)} \quad (9)$$

Then, the weighted hidden state $\tilde{h}_t^s$ and the context vector are defined as follows:

$$\tilde{h}_t^s = (\gamma_t^1 h_1^s, \gamma_t^2 h_2^s, \dots, \gamma_T^i h_T^s)^T \quad (10)$$

$$c_t = \sum_{j=1}^{T} \gamma_t^j h_j^s \quad (11)$$

Where $v_d, b_d \in R^p$, $W_d \in R^{q \times 2p}$, $U_d \in R^{p \times p}$ are parameters to learn, and $h_{t-1}^o \in R^p$ and $s_{t-1}^o \in R^p$ are respectively the hidden state and cell state of the previous hidden LSTM unit in the decoder, and $p$ is the hidden size in this attention module. $h_i^s \in H^s$ represents the i-th encoder hidden state at the second (or last) stage attention module. The context vector $c_t$ means the fuse information of all hidden state in the encoder, which represents the temporal relationships within a time window.

**4.3 Encoder-decoder & Model Training**

In both the first phase attention and second phase attention, we use LSTM unit of one layer to encode all series into the feature representation of the hidden state. The aim is to learn the following feature representations from input data $x_t$ at time *t*:

$$h_t = f_e(h_{t-1}, x_t) \quad (12)$$

Where $h_t \in R^m$ is the hidden state of the first phase attention, *m* is the size of the hidden state in the first phase, and $f_e$ is an LSTM unit. Note that all attention modules follow the above basic mapping in both the two-phase spatial attention stage and temporal attention stage. The difference is the different input series and the independent LSTM units, i.e., $f_e$ in Eqn. (11) and $f_d$ in Eqn. (13), are independent. The flow of the specific input series and output series is noted in Figure 1.

In the decoder, once we get the weighted summed context vector $c_t$, we combine the context vector $c_t$ and the target series **Y** at corresponding time:

$$\tilde{y}_t = \tilde{w}^T [y_t; c_t] + \tilde{b} \quad (13)$$

Where $\tilde{w}^T \in R^{q+1}$ and $\tilde{b} \in R$ are the parameters that map the concatenation to the size of the decoder input. Aligning the target series with context vectors makes it easier to maintain temporal relationships, and the result can be used to update the decoder hidden state:

$$d_t = f_d(d_{t-1}, \tilde{y}_{t-1}) \quad (14)$$

Where $d_t \in R^p$ is the hidden state used in the decoder, and $f_d$ is also an LSTM unit. Finally, we concatenate the context vector $c_t$ with the hidden state $d_t$, which is used as the new hidden state to make the final multi-step predictions:

$$\hat{y}_T, \hat{y}_{T+1}, \ldots, \hat{y}_{T+\tau} = v_y^T (W_y[d_t; c_t] + b_y) + b'_y \quad (15)$$

Where $W_y \in R^{p*(p+q)}$ and $b_y \in R^p$ map the concatenation $[d_t; c_t] \in R^{p+q}$ to the size of the decoder hidden state. The linear function with weights $v_y \in R^{\tau*p}$ and bias $b'_y \in R^\tau$ generates the final prediction result.

Since all models proposed in the paper are smooth and differentiable, we employ back-propagation algorithm to train all models. During the training process, we use minibatch stochastic gradient descent (SGD) together with the Adam optimizer to minimizing the mean squared error (MSE) between the predicted vector $\hat{y} \in R^\tau$ and the ground truth vector $y \in R^\tau$:

$$L(\theta) = \|\hat{y} - y\|_2^2 \quad (16)$$

**5 Experiments**

We implement all proposed models and baseline methods in PyTorch framework. In this section, we first describe different datasets from four fields and the introduction of baseline methods. Then, we introduce the parameter setting and model evaluation methods. Finally, extensive experiments have proved the effectiveness of our models. Moreover, we compare the effects of each module on experimental results, and we also provide an interpretation of attention-based RNN including our proposed models.

## 5.1 Datasets & Baseline methods

In order to demonstrate the generalization ability, we use four open datasets in the fields of energy, finance, environment and medicine to test our methods. The dataset information is shown in Table 1. We roughly divide the training set and test set into 4:1. In SML2010 dataset, we allocate training set and test set into about 3:1 because of the small amount of raw data. To demonstrate the effectiveness of our methods, we compare nine baseline models, e.g., classical non-linear prediction methods, kernel methods and state-of-the-art neural networks. We have compared the prediction results in the same time step size to evaluate the performance of all methods. We describe the datasets and baseline methods in detail as follows.

Table 1 Information of the dataset

| Dataset | Abbreviation | Field | Exogenous series | Size | |
|---|---|---|---|---|---|
| | | | | Train | Test |
| [1]SML2010 Data Set | SML2010 | Environment | 18 | 2000 | 763 |
| [2]NASDAQ 100 stock data | NASDAQ100 | Finance | 81 | 32000 | 8560 |
| [3]Appliances energy prediction Data Set | Energy | Energy | 27 | 16000 | 3736 |
| [4]EEG Steady-State Visual Evoked Potential Signals Data Set | EEG | Medicine | 13 | 10000 | 2288 |

### 5.1.1 Datasets

**SML2010:** This dataset is collected for indoor temperature prediction from a monitor system mounted in a domestic house. The data was sampled every minute, computing and uploading it smoothed with 15 minute means. It corresponds to approximately 40 days of monitoring data and is divided into two folders, but we only use the first folder, including about 30 days, as the whole dataset. In our experiment, we employ the room temperature as the target series and select 18 relevant exogenous series, while delete the date, time and three on-off series which are format-inconsistent attributes in the raw dataset. We use the first 2000 data points as the training set and the rest 763 data points as the test set.

**NASDAQ100:** This dataset is collected minute by minute under NASDAQ 100 for time series prediction and stock market analysis, which includes 105 days' stock data starting from July 26, 2016 to December 22, 2016. Each day contains 390 data points except for 210 data points on November 25 and 180 data points on December 22. There are in total 81 major corporations in this dataset and the missing data is interpolated with linear interpolation, which is all used as exogenous series in this study. The index value of the NASDAQ 100 is used as the target series. In our experiment, we employ the first 32000 data points as the training set and the rest 8560 data points as the test set.

**Energy:** This dataset can be employed to predict appliances energy use in a low energy building. The dataset is at 10 minute for about 4.5 months. In our experiment, we employ appliances energy use as target series, delete date attribute, and employ other attributes as exogenous series. We use the first 16000 data points as the training set and the rest 3736 data points as the test set.

**EEG:** This dataset consists of 30 subjects performing Brain Computer Interface for Steady State Visual

---

[1] https://archive.ics.uci.edu/ml/datasets/SML2010
[2] https://cseweb.ucsd.edu/~yaq007/NASDAQ100_stock_data.html
[3] https://archive.ics.uci.edu/ml/datasets/Appliances+energy+prediction
[4] https://archive.ics.uci.edu/ml/datasets/EEG+Steady-State+Visual+Evoked+Potential+Signals

Evoked Potentials (BCI-SSVEP), and we only use the visual image search dataset from the first subject. For visual experiments, since the electrodes O1 and O2 are more important information, we employ the electrodes O1 attribute as target series. Visual response prediction provides a reference for behavioral analysis and emotional orientation. In our experiment, we use other 13 signals attributes coming from the electrodes as exogenous series. Finally, we use the first 10000 data points as the training set and the rest 2288 data points as the test set.

**5.1.2 Baseline methods**

**ARIMA:** It is a well-known method for time series prediction, which is a development of the autoregressive moving average (ARMA) model(Amini, et al., 2016).

**SVR:** The kernel method based on support vectors has performed well in time series prediction so far, and the computer science community is still looking for loss representation methods to represent inter-class intervals of datasets with different distribution characteristics. In this paper, we employ SVR method with linear, polynomial and radial basis function kernel respectively, and then select the best results(Jie & Zio, 2016).

**LSTM:** This method can overcome the limitation of vanishing gradient in RNNs, and thus can capture long-term dependence in sequential learning(Graves, 1997).

**GRU:** Compared with LSTM, GRU has one less gating unit, which leads to fewer parameters and easy convergence, but the former performs better when the dataset is big(Chung, et al., 2014).

**Encoder-Decoder:** This method is widely used in machine translation, which encodes the source sequence into a fixed-length vector and generates the translation using the decoder (Cho, Merrienboer, et al., 2014).

**Input-Attn-RNN:** Based on encoder-decoder method, we only employ attention mechanism on the raw dataset and the others remain unchanged(Qin, et al., 2017).

**Temp-Attn-RNN:** Based on encoder-decoder method, we only employ attention mechanism on the hidden states and the others remain unchanged. This is a method that we propose for ablation experiments to test the effect of only using temporal attention module on experiment performance.

**DA-RNN:** This method is one of the state-of-the-art methods for single-step time series prediction. Specifically, the attention mechanism is used in both raw dataset and hidden states. The former extracts spatial correlations, while the latter guarantees long-term dependence(Qin, et al., 2017).

**GeoMAN:** The basic structure of this method is very similar to DARNN, and it is a state-of-the-art method for geosensory time series prediction. Since there are no global data from different locations and the external factors, we delete the global spatial attention and concatenate the past values of the target series directly into the spatial attention instead of the external factors(Yuxuan, et al., 2018).

**5.2 Parameter setting & Evaluation metrics**

For long-term prediction, the time step $\tau$ of the previous works is short, e.g., many advanced neuron network-based works(Liu, Zheng, Liang, Liu, & Rosenblum, 2016; Yu, et al., 2015; Yuxuan, et al., 2018) only set $\tau = 6$. To demonstrate that our method is more suitable for long-term prediction, we select the time step $\tau \in \{5,10,30,50,120\}$. During the training process, the batch size is set to 128, and the learning rate is set to 0.001. Totally, there are two types of key hyperparameters in our model, i.e., the size of time window $T$ and the size of hidden states for each attention module. We conducted a grid search and select the best performance for the former because different time steps perform well in different time windows, and we observe that the best result can be achieved in most cased when $T$ is equal to 5 or 10. For simplicity, we employ LSTM network of one layer and the same size of hidden states for each attention module, i.e., m = p = q, and also conducted a grid search to achieve the best results. The setting in which m = p = q = 128 outperforms the others in most cases and we set it to this fixed value for all models, and the value is also used for DeepAttn model. For the reproducibility of experiment results, we set random seeds to 2019 in all experiments.

Follow the previous work(Yuxuan, et al., 2018), we employ root mean squared error (RMSE) and mean absolute error (MAE) to evaluate our models, both of which are scale-dependent and widely used in time series prediction. Specifically, assuming $y_t$ is the ground truth and $\hat{y}_t$ is the predicted value at time $t$, they are defined as follows:

$$\text{RMSE} = \sqrt{\frac{1}{N}\sum_{i=1}^{N}(\hat{y}_t^i - y_t^i)^2} \qquad (17)$$

$$MAE = \frac{1}{N}\sum_{i=1}^{N}|\hat{y}_t^i - y_t^i| \qquad (18)$$

**5.3 Results & Discussion**

In this section, we evaluate the effectiveness of our methods from many perspectives. First, we compared our method with nine baseline methods on four datasets. Then, we give the functional interpretation to each module in our models, and find the negative performance of the deep spatial attention model and then explain the reasons. Next, we investigate ablation experiments on each module. Finally, we study the performance of our methods in different time steps.

**5.3.1 Model comparison**

In this part, we compared our methods with nine baseline methods on four different datasets to prove the effectiveness of our models. To be fair, we show the best results of each method under different parameter settings. The experimental results of forecasting 30 time steps and 120 time steps are shown in Tables 2 and Table 3, which represent the relatively short-term and relatively long-term forecasting, respectively.

In both Tables 2 and Table 3, the DSTP-RNN and DSTP-RNN-Ⅱ model exceed the state-of-the-art models in all datasets. We also observed that DSTP-RNN-Ⅱ achieves better results than DSTP-RNN when the time step is short, because a combination of multiple attention modules for all series can capture more tiny spatial correlations. Conversely, DSTP-RNN performs better when the time step is longer. This is because that directly focuses on the same spatial correlations can learn more stable long-term dependence, and these high-level spatial features is not susceptibility to interference for long-term prediction.

Table 2: Performance comparison among different methods and datasets when the time step $\tau$ is equal to 120 (first line represents RMSE, second line represents MAE, and best result displays in **boldface**)

| Method | SML2010 | NASDAQ100 | Energy | EEG |
|---|---|---|---|---|
| ARIMA | 1.0729 | 0.9397 | 0.8564 | 1.2311 |
|  | 0.9066 | 0.7400 | 0.5026 | 0.9930 |
| SVR | 0.6049 | 1.0446 | 0.8589 | 1.3393 |
|  | 0.4751 | 0.8557 | 0.8185 | 0.9688 |
| LSTM | 0.7561 | 0.8965 | 0.3682 | 1.4959 |
|  | 0.5663 | 0.6935 | 0.5545 | 1.1969 |
| GRU | 0.7770 | 1.5762 | 0.8671 | 1.1521 |
|  | 0.5829 | 1.3679 | 0.5042 | 1.2210 |
| Encoder-Decoder | 0.5875 | 1.1288 | 0.8803 | 0.9155 |
|  | 0.4091 | 0.8678 | 0.4172 | 0.7142 |
| Input-Attn-RNN | 0.4221 | 0.8193 | 0.8706 | 0.7689 |
|  | 0.2694 | 0.6609 | 0.4792 | 0.6153 |
| Temp-Attn-RNN | 0.3813 | 0.3513 | 0.7548 | 0.8211 |
|  | 0.2942 | 0.2927 | 0.3579 | 0.6452 |
| DARNN | 0.4180 | 0.1541 | 0.7656 | 0.7084 |
|  | 0.2714 | 0.1203 | 0.3631 | 0.5470 |
| GeoMAN | 0.4272 | 0.3341 | 0.7882 | 0.7076 |
|  | 0.3077 | 0.2352 | 0.3795 | 0.5303 |
| DeepAttn | 0.3368 | 0.1418 | 0.7952 | 0.8089 |
|  | 0.1873 | 0.1003 | 0.3817 | 0.5805 |
| DSTP-RNN | **0.2962** | **0.0895** | **0.7477** | **0.6535** |
|  | 0.1597 | **0.0661** | **0.3431** | **0.4996** |
| DSTP-RNN-II | 0.3087 | 0.1214 | 0.7524 | 0.6608 |
|  | **0.1593** | 0.0739 | 0.3528 | 0.5059 |

Table 3: Performance comparison among different methods and datasets when the time step $\tau$ is equal to 30 (first line represents RMSE, second line represents MAE, and best result displays in **boldface**)

| Method | SML2010 | NASDAQ100 | Energy | EEG |
|---|---|---|---|---|
| ARIMA | 1.0631 | 0.9530 | 0.8491 | 1.2022 |
|  | 0.7980 | 0.7846 | 0.4994 | 0.9618 |
| SVR | 0.6843 | 0.2821 | 0.8336 | 1.0464 |
|  | 0.7788 | 0.2324 | 0.4439 | 0.7664 |
| LSTM | 0.7016 | 0.9592 | 0.8368 | 1.1965 |
|  | 0.5320 | 0.7424 | 0.4076 | 0.9634 |
| GRU | 0.7084 | 0.6649 | 0.9071 | 1.1701 |
|  | 0.5405 | 0.5131 | 0.4531 | 0.9395 |
| Encoder-Decoder | 0.2537 | 0.3956 | 0.8039 | 1.1854 |
|  | 0.2050 | 0.2922 | 0.3908 | 0.9623 |
| Input-Attn-RNN | 0.2144 | 0.2642 | 0.7982 | 0.9846 |
|  | 0.1632 | 0.2297 | 0.3874 | 0.7454 |
| Temp-Attn-RNN | 0.2406 | 0.3515 | 0.7880 | 0.8104 |
|  | 0.1900 | 0.2992 | 0.4271 | 0.6363 |
| DARNN | 0.2080 | 0.1958 | 0.7551 | 0.7637 |
|  | 0.1693 | 0.1661 | 0.3700 | 0.5616 |
| GeoMAN | 0.1310 | 0.2304 | 0.7847 | 0.7439 |
|  | 0.1022 | 0.1780 | 0.4024 | 0.5653 |
| DeepAttn | 0.1647 | 0.2439 | 0.8089 | 0.8248 |
|  | 0.1206 | 0.1925 | 0.5805 | 0.5478 |
| DSTP-RNN | **0.0987** | 0.0987 | 0.7792 | 0.6504 |
|  | **0.0761** | 0.0761 | 0.3703 | 0.4969 |
| DSTP-RNN-Ⅱ | 0.0993 | **0.0661** | **0.7706** | **0.6403** |
|  | 0.0795 | **0.0507** | **0.3296** | **0.4864** |

In detail, the performance of ARIMIA is much worse than other methods, which indicates that this method that only employs target series while ignoring exogenous factors is not suitable for multivariate time series prediction. The SVR model based on kernel methods only captures discrete spatial correlations, while the single LSTM and GRU model pays more attention to maintaining time dependence. These two methods have different performances in different datasets, and the average performance is comparable. The number of exogenous attributes in datasets and the size of training data are the reasons for the performance difference between these two methods. The spatial correlations of the former are more complicated, and the latter is more suitable for the methods which can maintain time dependence. The encoder-decoder model generally performs better than the single LSTM and GRU model, but is affected by different datasets, so the fixed-length vector is not sufficient to represent the long-term dependence characteristics. However, the spatial attention mechanism for different series and the temporal attention mechanism for the time dimension can improve the performance of encoder-decoder model, indicating that weighted representation of raw data in a single spatial or temporal dimension can also improve the forecasting performance.

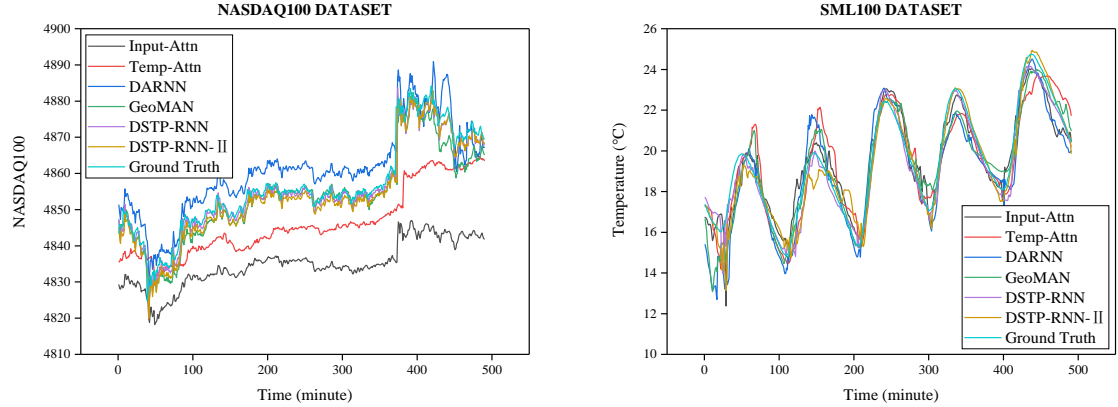

Figure 2: Prediction results of different methods on NASDAQ100 dataset and SML2010 dataset ($\tau = 120$).

The DARNN method is the state-of-the-art method in single-step prediction, and GeoMAN is the most advanced method for geosensory time series. The performance of these two methods in long-term prediction is also better than single-stage attention mechanism. Nevertheless, both DSTP-RNN and DSTP-RNN-II outperform these methods in all datasets. Since the weight change from attention mechanism of the single phase is violent but decentralized, it is not enough to accurately and stably represent spatial correlations. In contrast, the weight distribution of the second phase attention is more stationary and concentrated, thus learning more stable spatial correlations. Besides, it ensures time dependence more effectively in the second phase attention by concatenating target series to obtain spatio-temporal relationships between exogenous series and target series. To visualize the prediction results, we present experimental results of six models in Figure 2, which is based on the test set of NASDAQ100 dataset and SML2010 dataset when the time step is set to 120. We can observe that prediction results of our models are very close to the ground truth when the training data is sufficient in NASDAQ100 dataset, and our methods also outperform other baseline methods when the training data is insufficient in SML2010 dataset. To make the data change clearer, we only show a comparison of 500 prediction values and real values. In addition, since prediction values of the single LSTM and GRU model and other baseline models in long-term prediction have large errors against the true values, we omit them in Figure 2.

### 5.3.2 Model interpretation & Deep spatial attention mechanism

In the attention-based RNN models, the spatial correlation is determined by assigning weights to the different attributes, and the temporal relationship is captured by establishing a temporal attention mechanism for encoder hidden states. The spatio-temporal relationship is determined on the one hand by backpropagation of the connective attention structures, and on the other hand by encoding all series within a fixed time window.

The proposed models inherit the above advantages. In addition, we employ a two-phase spatial attention mechanism to obtain more accurate and stable spatial correlations, in which the weight generated by the first phase is violent but decentralized, and the weight generated by the second phase is concentrated on the most relevant exogenous series. Figure 3 shows the weight distribution from a single encoder time step based on the three-layer spatial attention mechanism, i.e., DeepAttn. The encoder of other time steps reflects similar feature distributions. We can observe that in addition to these most relevant exogenous attributes, the weight distribution of other attributes is very stationary in the second layer.

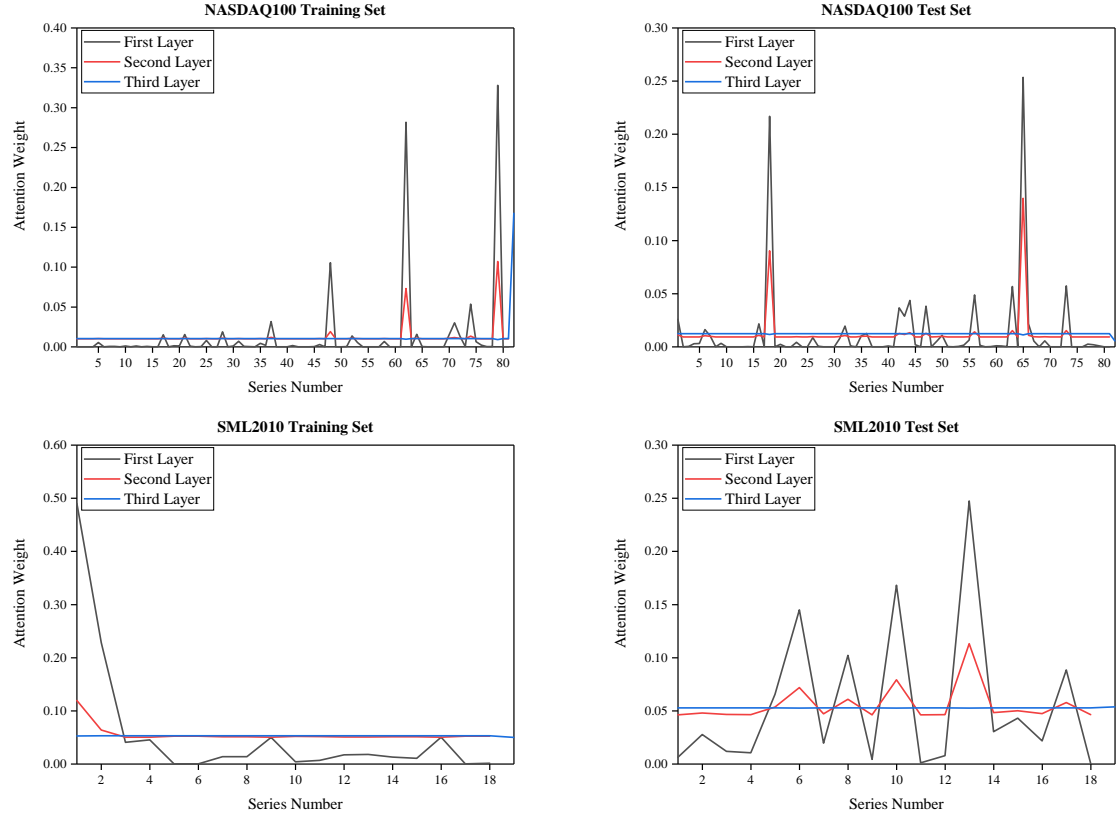

Figure 3: Distribution of attention weights in different spatial phase from a single encoder time step on train set and test set of NASDAQ100 dataset and SML2010 dataset.

Since third-layer attention weights are averaged distribution in Figure 3, this layer and the deeper layers do not work at all. However, it is obvious in Table 1 that the third-layer spatial attention mechanism (i.e., DeepAttn) does not enhance the performance, but it has a negative impact on the experimental results. This is because although the third-layer attention assigns the average weights to all exogenous attributes, the convergence performance of the model may even worse when the network is deeper (Allen-Zhu, Li, & Song, 2018; Arora, Basu, Mianjy, & Mukherjee, 2018). Still, the performance of DeepAttn outperforms DARNN and GeoMAN.

In Figure 3, note that the last attention weight of the third layer is from the target attribute, which represents the spatial correlations between target attributes and exogenous attributes. Therefore, another advantage of DSTP-RNN and DSTP-RNN-Ⅱ is that we not only use the temporal attention mechanism to select relevant hidden states in all time windows, but also concatenate the target attribute corresponding to the exogenous attribute in the second phase attention, thereby increasing spatio-temporal relationships between target series and exogenous series. In addition, the DSTP-RNN-Ⅱ structure is inspired by the influence of target information and non-target information in human attention mechanism, which can better imitate the feature extraction structure of human neural networks. This method works best in relatively short-term prediction, and the results in long-term prediction also outperform all baseline methods.

**5.3.3 Ablation experiment**

We construct a two-phase spatial attention mechanism in the model structure and give attention weight to the target series in the second phase attention. Also, we develop the DSTP-RNN-Ⅱ model according to the biphasic effect mechanism of target information and non-target information in human attention model. Accordingly, we mainly compare the effect of different attention modules and the biphasic effect mechanism with or without the target information on experimental results.

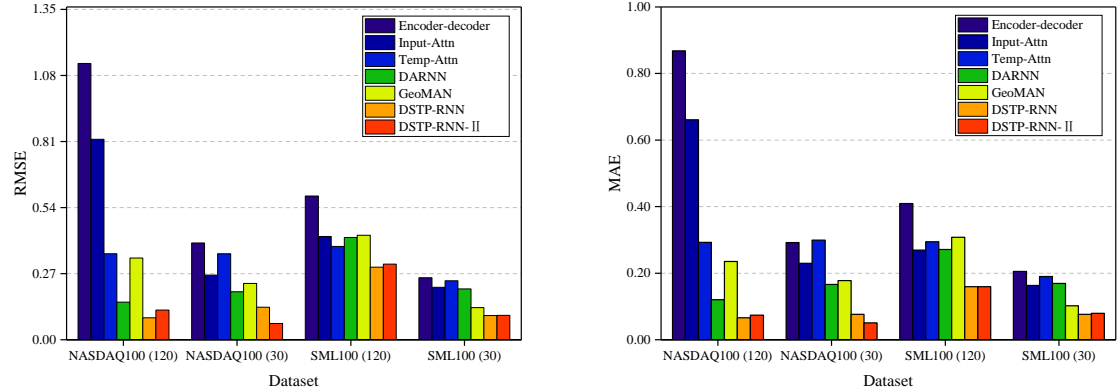

Figure 4: Performance of different methods based on different attention structure. The results of NASDAQ100 dataset and SML2010 dataset in different time step prediction are presented, e.g., NASDAQ100 (120) means that the time step is set to 120.

In Figure 4, we can observe that the laws in all datasets are roughly the same: with the development of encoder-decoder model with no attention mechanism and the other models with one or more attention mechanism, the model continues to achieve better results. The performance of DARNN and GeoMAN are superior to a single attention model in almost all datasets and different time steps, and it only has no significant effect on in SML2010 dataset. Obviously, the effect of the two-phase attention-based structure is superior to the state-of-the-art models in the different time step of all datasets. We can also observe that the DSTP-RNN-Ⅱ model closer to the human attention mechanism can achieve better results in the short-term prediction, which indicates that both target information and non-target information play an important role in capturing more detailed spatial correlations. Figure 5 represents the impact of our models with or without combining past target information to the last phase attention on experimental results. We find that the concatenation with target series information in second (or last) phase attention can better maintain the time dependence between target series and exogenous series. All experimental results validate the effectiveness of different modules in our models.

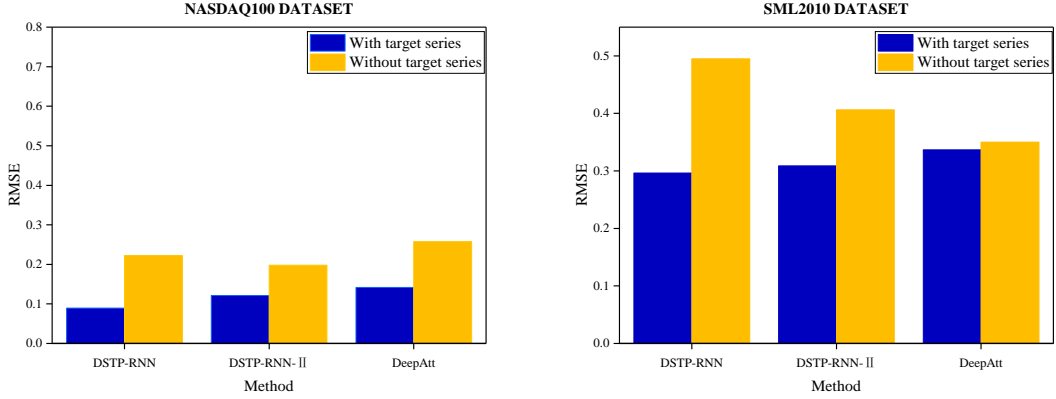

Figure 5: Effect of target information mechanism on NASDAQ100 dataset and SML2010 dataset.

**5.3.4 Long-term prediction analysis**

Long-term prediction is more significant in practical applications, which can provide sufficient buffer time for the next action. Figure 6 shows the experimental results in different time step prediction. We can observe that as the time step $\tau$ increases, the prediction performance will gradually decrease. It is also found from the NASDAQ100 dataset that this trend is relatively slow because the amount of training data is sufficient. In addition, the DSTP-RNN model and DSTP-RNN-Ⅱ model outperform the state-of-the art methods in both short-term and long-term prediction. Furthermore, the performance of DSTP-RNN-Ⅱ model in short-term prediction is better than DSTP-RNN model in most instances.

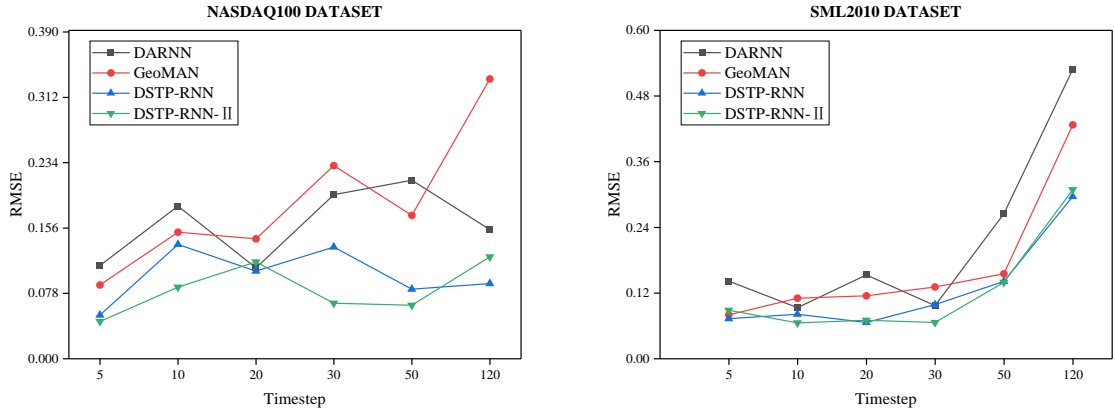

Figure 6: Performance of different time step prediciton based on NASDAQ100 dataset and SML2010 dataset. The state-of-the-art methods are compared with our proposed methods.

**6 Conclusion & Future work**

In this paper, we propose two novel attention-based RNN for long-term and multivariate time series prediction, i.e., DSTP-RNN and DSTP-RNN-Ⅱ. In general, our models enhance the attention mechanism of both spatial correlations and temporal relationships to better capture spatio-temporal relationships, and thus outperform the state-of-the-art methods in four datasets and different time step prediction. Our interpretation of the attention-based model provide a developed idea for further understanding the spatio-temporal relationships of time series, and for further exploring the attention-based methods in time series prediction. Extensive experiments have proved the following conclusions:

- The DSTP-RNN and DSTP-RNN-Ⅱ model have achieved the best results in the long-term prediction, indicating that the bionic structure of artificial neural network has good performance in practical applications. Similar to the neural response in human DSTP model, the attention weight of the first phase is violent but decentralized, and the second phase leads to stationary and concentrated response weight, so the extracted spatial correlations are more stable and effective.

- Due to the effectiveness both of target information and non-target information in the human neuron signal, the DSTP-RNN-Ⅱ model and the method of concatenating the corresponding target information in the last phase can effectively improve the ability of the model to capture spatio-temporal relationships between target series and exogenous series.
- A deeper spatial mechanism is not necessary since the attention weights in the deeper attention layer are evenly distributed. Also, the prediction performance decreases as the time step increases.

In the future, the key issue in achieving accurate long-term prediction of time series is to solve the problem: how to represent the three relationships that is presented at the beginning of this paper. The attention-based RNN is a good direction, and there are two trends in the development of these methods for time series prediction. On the one hand, the attention mechanism can be combined with the convolutional neural network (CNN) module. Since CNN is more sensitive to spatial information, the attention module can become the connection module or a high-level feature selection module between CNN and RNN. On the other hand, it is important to use the attention mechanism to simultaneously learn spatio-temporal attention weights from different spatial and temporal dimensions.